\documentclass[sigconf,screen]{acmart}
\usepackage{multirow}
\AtBeginDocument{%
  }

\copyrightyear{2024}
\acmYear{2024}
\setcopyright{cc}
\setcopyright{cc}
\setcctype{CC-BY}
\acmConference[VRCAI '24]{The 19th ACM SIGGRAPH International Conference
on Virtual-Reality Continuum and its Applications in Industry}{December 1--2,
2024}{Nanjing, China}
\acmBooktitle{The 19th ACM SIGGRAPH International Conference on
Virtual-Reality Continuum and its Applications in Industry (VRCAI '24), December
1--2, 2024, Nanjing, China}\acmDOI{10.1145/3703619.3706036}
\acmISBN{979-8-4007-1348-4/24/12}
\setcopyright{acmlicensed}

\acmSubmissionID{1556}

\begin{document}

\newcommand{\yq}[1]{{\color[rgb]{0.2,0.6,0.6}(Yongqing: #1)}}

\title{AURORA: Automated Unleash of 3D Room Outlines for VR Applications}

\author{Huijun Han}

\email{hazelhan@tamu.edu}
\orcid{0009-0009-1360-4886}

\affiliation{%
  \institution{Texas A\&M University }
  \city{College Station}
  \state{Texas}
  \country{USA}
}

\author{Yongqing Liang}
\email{lyq@tamu.edu}
\orcid{0000-0002-7282-0476}

\affiliation{%
  \institution{Texas A\&M University}
  \city{College Station}
  \state{Texas}
  \country{USA}
}
\author{Yuanlong Zhou}
\email{ryanbowz@tamu.edu}
\orcid{0009-0002-4286-685X}

\affiliation{%
  \institution{Texas A\&M University}
  \city{College Station}
  \state{Texas}
  \country{USA}
}

\author{Wenping Wang}
\email{wenping@tamu.edu}
\orcid{0000-0002-2284-3952}

\affiliation{%
  \institution{Texas A\&M University}
  \city{College Station}
  \state{Texas}
  \country{USA}
}

\author{Edgar J. Rojas-Muñoz}
\email{ed.rojas@tamu.edu}
\orcid{0000-0001-6909-375X}

\affiliation{%
  \institution{Texas A\&M University}
  \city{College Station}
  \state{Texas}
  \country{USA}
}
\author{Xin Li}
\email{xinli@tamu.edu}
\orcid{0000-0002-0144-9489}

\affiliation{%
  \institution{Texas A\&M University}
  \city{College Station}
  \state{Texas}
  \country{USA}
}

\renewcommand{\shortauthors}{Han et al.}

\begin{abstract}
Creating realistic VR experiences is challenging due to the labor-intensive process of accurately replicating real-world details into virtual scenes, highlighting the need for automated methods that maintain spatial accuracy and provide design flexibility.
In this paper, we propose AURORA, a novel method that leverages RGB-D images to automatically generate both purely virtual reality (VR) scenes and VR scenes combined with real-world elements.
This approach can benefit designers by streamlining the process of converting real-world details into virtual scenes.
AURORA integrates advanced techniques in image processing, segmentation, and 3D reconstruction to efficiently create realistic and detailed interior designs from real-world environments.
The design of this integration ensures optimal performance and precision, addressing key challenges in automated indoor design generation by uniquely combining and leveraging the strengths of foundation models.
We demonstrate the effectiveness of our approach through experiments, both on self-captured data and public datasets, showcasing its potential to enhance virtual reality (VR) applications by providing interior designs that conform to real-world positioning.
\end{abstract}

\begin{CCSXML}
<ccs2012>
   <concept>
       <concept_id>10010147.10010178.10010224.10010245.10010254</concept_id>
       <concept_desc>Computing methodologies~Reconstruction</concept_desc>
       <concept_significance>500</concept_significance>
       </concept>
   <concept>
       <concept_id>10010147.10010178.10010224.10010225.10010227</concept_id>
       <concept_desc>Computing methodologies~Scene understanding</concept_desc>
       <concept_significance>500</concept_significance>
       </concept>
   <concept>
       <concept_id>10003120.10003121.10003124.10010866</concept_id>
       <concept_desc>Human-centered computing~Virtual reality</concept_desc>
       <concept_significance>500</concept_significance>
       </concept>
 </ccs2012>
\end{CCSXML}

\ccsdesc[500]{Computing methodologies~Reconstruction}
\ccsdesc[500]{Computing methodologies~Scene understanding}
\ccsdesc[500]{Human-centered computing~Virtual reality}

\keywords{Virtual Reality,  Interior Design, Room Layout}

\begin{teaserfigure}
  \includegraphics[width=\textwidth]{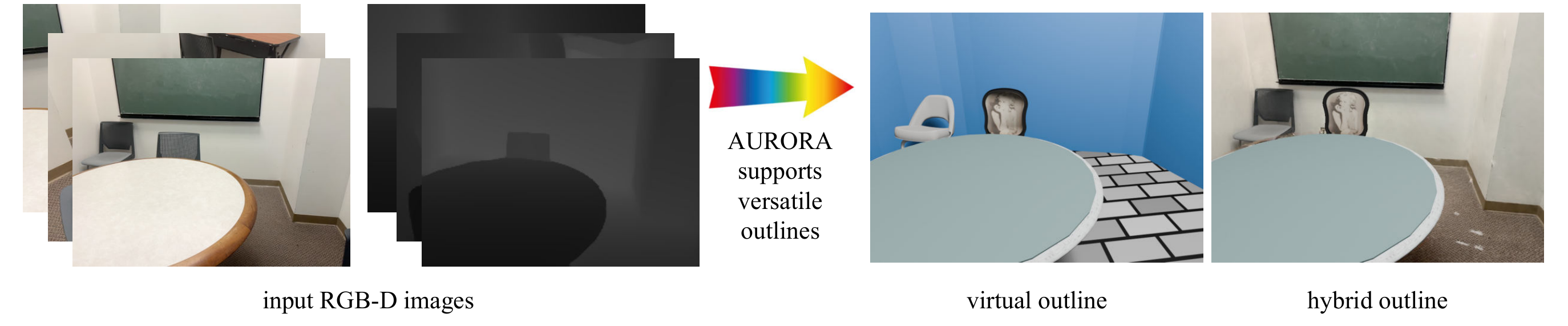}
  \caption{This figure illustrates the capability of AURORA to generate multiple outlines from input RGB-D images. The virtual outlines represent a fully modeled 3D scene with complete object replacements, while the hybrid outlines combine both the reconstructed geometry and model placements, providing a comprehensive yet flexible visualization of the interior design layout. }
  \label{fig:teaser}
\end{teaserfigure}

\maketitle
\section{Introduction}

% background 
Creating Virtual Reality (VR) is essential for delivering engaging user experiences.  However, building high-quality virtual scenes is often labor intensive and time-consuming. 
Designers often need to invest substantial expertise and effort to meticulously translate real-world details into virtual settings, ensuring accuracy, engagement, and visual coherence.
To infuse authenticity into VR spaces, designers often draw inspiration from their own surroundings, modeling virtual scenes after real-life environments or incorporating structural and aesthetic cues from the physical world. This approach brings familiarity to VR experiences, making them more relatable and easier to navigate. 
However, this  manual process of capturing and replicating these details is inefficient.
Thus, it is highly desirable to have efficient and automatic methods to seamlessly convert real-world details into virtual scenes populated with 3D assets in correct orientation, dimension, and layout.

Besides automation, designers often seek flexibility in the design process. For example, even with a same layout design, different users could prefer using different material or styles of objects. Thus, it is desirable to have the capability of exploiting different design/style options from a same set of input images. 
A system that can use a same set of input images, to preserve the objects' spatial arrangement, yet generate 
both purely virtual reality (VR) environments and VR combined with real-world elements
will offer such a desirable flexibility requirement. 
In this work, we focus on indoor scene design generation.

Compared to directly using fisheye or panoramic images, a 3D reconstructed scene offers a more accurate and realistic representation of the environment, enabling smoother, more intuitive interaction and minimizing distortion during navigation. Additionally, to achieve semantic understanding of the scene, the instance segmentation part is crucial.
Thus, we perform 3D \textbf{reconstruction} followed by \textbf{segmentation} and \textbf{registration} to accurately transform 2D images into detailed interior designs, ensuring precise object placement and spatial alignment.

Recent advancements in \textbf{reconstruction} have shown that Gaussian Splatting based algorithms offer superior quality in dense reconstruction, especially in novel view synthesis and achieving high levels of color realism~\cite{keetha:2024:CVPR:splatam, yan:2024:CVPR:gsslam, Matsuki:2024:CVPR:GSSLAM}. 
These algorithms are particularly effective at generating detailed 3D reconstructions from 2D images, capturing both spatial geometry and texture with impressive accuracy~\cite{10521791}.
However, as these algorithms are not specifically optimized for indoor scenes, the resulting 3D scans may exhibit some noise, such as floating point artifacts~\cite{sandstrom2024splat}.
This introduces the first challenge: the noise in the scan data can accumulate as it is processed through the pipeline, particularly when fed downstream to subsequent modules like segmentation and model replacement.
In contrast to well-annotated point cloud instance segmentation datasets, such as ScanNet++~\cite{yeshwanthliu2023scannetpp} and S3DIS~\cite{armeni2016s3dis}, which support robust model training, the \textbf{segmentation} module faces limitations due to the lack of a comprehensive GS-based segmentation annotation dataset. This scarcity hinders the robustness of Gaussian-based segmentation methods across varied conditions.
As a result, the segmentation process on Gaussians becomes more challenging.
While recent work has addressed some aspects of grouping~\cite{gaussian_grouping}, these efforts remain insufficient to produce bounding box layout, especially when dealing with complex structures and large rooms, since they lack precise boundary detection between individual instances. 
Large-scale environments introduce additional complexities such as increased data volume, higher computational demands, and greater variability in room layouts.
In the \textbf{registration} part of the pipeline, the lack of utilizing indoor layout priors poses a significant challenge. Without incorporating common design rules or architectural constraints, such as wall boundaries, furniture placement guidelines, and room proportions, the registration process may result in inaccuracies where furniture and objects appear to float or penetrate walls.

These aforementioned challenges can be summarized as follows:
\begin{itemize}
    \item The quality of 3D reconstruction may be insufficient for downstream tasks, as noisy data from the scanning process can lead to errors that accumulate throughout the pipeline, affecting the accuracy and quality of the final design.
    \item There is currently no end-to-end automated design pipeline that can generate complete designs directly from RGB-D data and output a hybrid of Gaussian and mesh representations.
    \item  Absence of room priors: The failure to integrate room layout priors into the module integration results in unrealistic furniture placement, such as floating objects or objects penetrating walls, during the registration 
\end{itemize}

In this work, we introduce AURORA -- an automated pipeline that streamlines interior design by efficiently capturing and translating real-world spaces into detailed 3D indoor outlines.

Specifically, we first perform a GS-based SLAM,
using SPLATAM~\cite{keetha:2024:CVPR:splatam}, 
from RGB-D data captured by portable devices such as smartphones.
We then perform surface reconstruction, incorporating two novel geometry losses to enhance accuracy and quality.
Next, to leverage robust Foundation models for 3D instance segmentation, we convert the 3D  Gaussians into a point cloud using TSDF-fusion~\cite{zeng20163dmatch}. 
Then, to ensure the segmented bounding box layout to be positioned on the ground, we enforce geometric constraints to (1) align the layout's base plane with the ground plane, and (2) preventing the layout from penetrating the walls or intersecting each others. 

Unlike common model replacement methods 
that often result in model overlap and orientation inconsistencies, our approach minimizes these issues and better preserves overall scene layout, maintaining consistency in both object placement and orientation. 

We conducted extensive experiments to validate our system, which can produce high-quality VR scenes from captured image sequences effectively. 
This flexibility and ease of use make it a useful tool that significantly enhances accessibility and efficiency in VR scene design, for not only designers but also common public users. 

The \textbf{main contributions} of this work are

% %\textcolor{purple}{
\begin{itemize} 
\item We refine the Gaussian representation with two novel geometry losses to ensure optimal performance in downstream tasks. 
\item We propose the first end-to-end pipeline that directly transforms RGB-D data into a mixed Gaussian and mesh representation.
\item We leverage interior priors to enhance the realism of the room layout, ensuring more accurate furniture placement and spatial arrangement.
\end{itemize}

\section{Related Work}

\subsection{3D Room Scanning for Virtual Reality}

In 3D room scanning, tools such as fisheye lenses, panoramic imaging, and RGB-D sensors like the iPhone with LiDAR scanner are frequently employed to capture and convert real-world surroundings into a virtual space.

Panoramic images, particularly 360-degree ones, are widely used for virtual reality room scanning due to their ability to provide an immersive environment by allowing users to explore real-world spaces~\cite{Cruz_2021_CVPR, shen2023disentangling, Tsai_2024_CVPR, zhou2025dreamscene360}.
However, there are several drawbacks when using 360$^\circ$ panoramas as a medium, such as limitations in image resolution, the fixed viewpoint which restricts movement within the scene, and parallax errors caused during the stitching process~\cite{ritter2022three}.

In virtual reality, fisheye lenses are beneficial as they capture extremely wide-angle images, thus enhancing the immersive experience by including more of the environment in a single shot~\cite{9583736,MENG2024102151}.
After capturing, the fisheye images need to be corrected for the distortion caused by the lens~\cite{amini2022vista}.
The primary issue with fisheye images is the inherent distortion. The barrel effect, which stretches the center of the image and compresses the periphery, can misrepresent the geometry of the space. 
For example, in a modeled camera translation, simulated images exhibit unrealistic perspective distortions~\cite{10462008}. In floor plans, this can be problematic when users need precise spatial information.

In contrast, the advantages of RGB-D sensors are three-fold.
First, it is affordable for most common users. Recent iPhones have powerful depth lenses to allow users to capture depth-of-field images.
Second, the RGB-D sensor is often portable and can be easily set in the field.
Third, images captured by the RGB-D sensor appear visually similar to what humans typically see, compared with the panoramic and fisheye lenses.
Hence, we leverage the RGB-D images as our 2D inputs to generate multiple interior designs.

\subsection{3D Instance Segmentation}

In the field of 3D instance segmentation, recent works primarily have two technical approaches: one is to segment based on 3D Gaussian Splatting (3D-GS), and another is to segment based on point cloud. 

Gaussian grouping~\cite{gaussian_grouping}
grouped items based on 2D segmentation labels by using Identity Encodings that link Gaussians to object instances, guided by 2D mask predictions from models like SAM during differentiable rendering.
SAGD~\cite{sagdboundary} addressed ambiguous boundaries in 3D-GS segmentations by using a Gaussian Decomposition scheme that learns from 3D-GS’s structure, improving boundary segmentation and accuracy.
However, due to the inherent complexity of 3D-GS, including its ambiguous structures and unconstrained geometry, these methods face challenges with unclear boundaries between objects and the background, leading to segmentation inaccuracies and reduced robustness.

On the other hand, point cloud segmentation yields more reliable results, as it benefits from a larger training dataset compared to 3D-GS-based segmentation.
Recently, SoftGroup++~\cite{vu2023scalable, vu2022softgroup} proposed a 3D instance segmentation model that can handle a wide variety of room types.
It is trained on the extensive ScanNet++ dataset~\cite{yeshwanthliu2023scannetpp}, enabling it to generalize well across various 3D scenes, ensuring broad applicability in real-world environments.
Another framework, named MSTA3D~\cite{10.1145/3664647.3680667}, addressed challenges such as over-segmentation and unreliable mask predictions in 3D instance segmentation. It achieves this by leveraging a multi-scale feature representation combined with a novel twin-attention mechanism, improving segmentation accuracy and robustness.

\subsection{3D Model Retrieval and Registration}

In \cite{ainetter2023automatically,10550876}, Ainetter presented an automatic method for aligning CAD models with captured scenes. To refine the pose of a CAD model, the method uses a differentiable pose refinement approach. The 9-DoF pose of the model is found using a differentiable optimization process that minimizes the error between the rendered CAD model and the captured scene. 
One main drawback of the differentiable pose refinement method is that it relies on accurate initial pose estimation. If the initial alignment is off, the refinement process might struggle to converge to a correct solution. 
Wei use learned representations to distinguish between model categories and a modified Chamfer distance metric for model registration, re-ranking the CAD neighborhood to enable fine-grained retrieval of clean CAD models from a large-scale database~\cite{wei2022accurate}.
Since these methods are not specifically designed for indoor furniture placement, they may not adequately address issues such as object intersection with walls and floors.

\section{Methodology}

\begin{figure*}[ht]
  \centering
  \includegraphics[width=\linewidth]{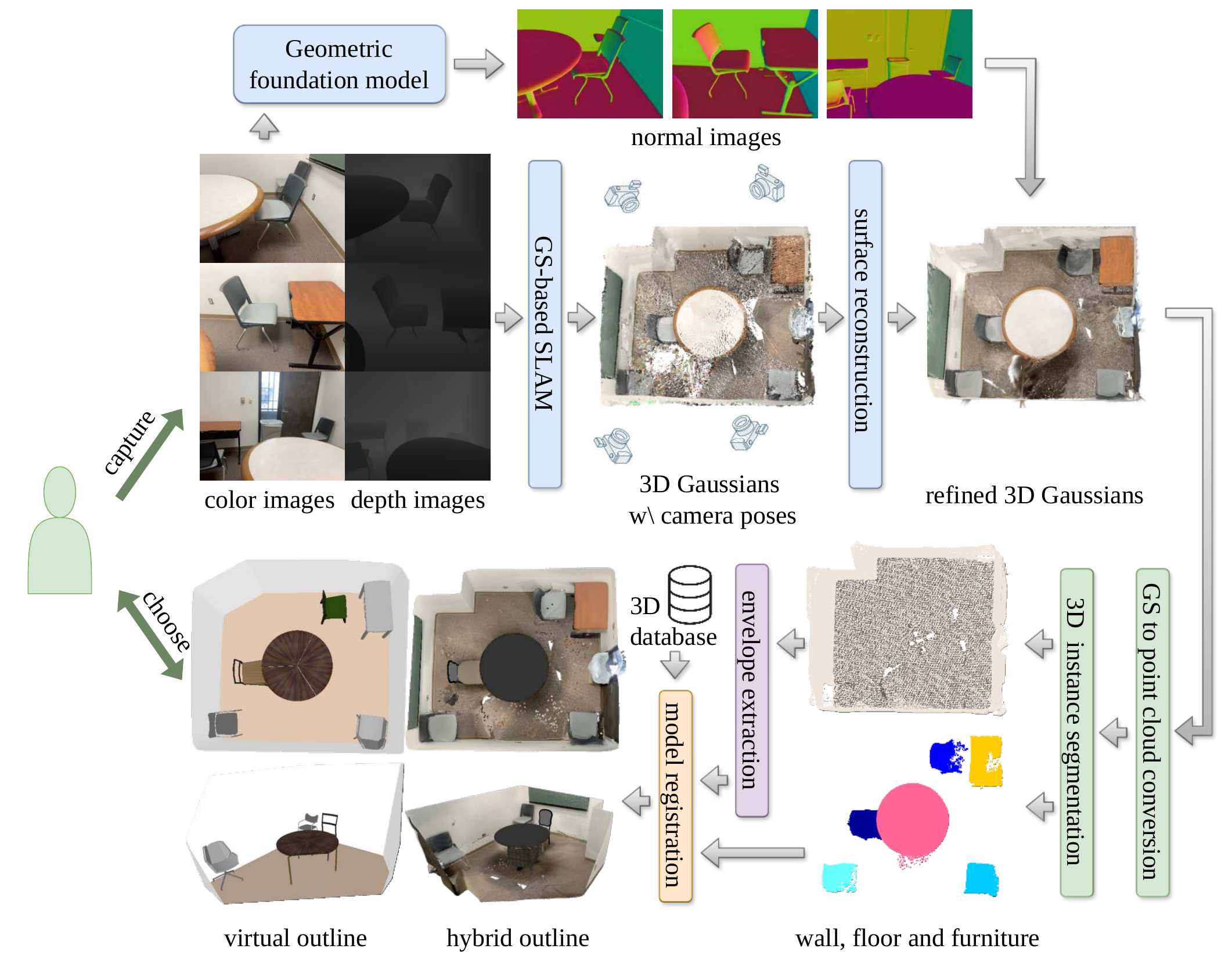}
  \caption{This illustration demonstrates our pipeline, where user-provided RGB-D images are processed to generate multiple room outline options. 
  The first option, named virtual outline, replaces the entire scene with CAD models for a fully structured representation. 
  The second option, named hybrid outline, replaces selected furniture items with CAD models while retaining other elements from the 3D reconstruction for a blended representation.
The input images are processed via a Gaussian Splatting-based module, generating 3D Gaussians with camera poses (top middle), then refined with normal priors to improve accuracy and detail  (top right).
This step is followed by a conversion from GS to a textured point cloud for feeding into the instance segmentation module.
The 3D segmentation module then identifies and separates walls, floors, and furniture.
Segmented walls and floors (bottom middle) are used to extract an actual-size floor with surrounding walls, while segmented furniture (bottom middle) is matched to database models, presenting outline options to the user.
}
\label{figure:pipeline}
\end{figure*}
We propose an automated pipeline that takes user-provided RGB-D images as input and generates room outline options---either virtual or hybrid---as output, as shown in Figure~\ref{figure:pipeline}.

The virtual outline option, where the entire environment is replaced with CAD models, ensures a fully structured representation beneficial for applications requiring precision and standardization, such as interior design and virtual staging. 
The hybrid outline option, which replaces only part of the furniture with CAD models while retaining reconstructed 3D Gaussians, offers a balance between realism and flexibility, making it suitable for scenarios like renovation planning or mixed-reality experiences.

The methodology is detailed as follows: Section~\ref{SLAM} covers the 3D reconstruction process the necessary refinement steps to enhance the accuracy and detail of the reconstructed scene.
Section~\ref{softgroup} describes the instance segmentation module, which identifies and segments walls, floors, and furniture, including the necessary conversion of the map representation. 
Section~\ref{dim-est} estimates the floor and wall objects to compute the indoor dimension.
Section~\ref{regis} outlines the model registration step, which produces the interior room design options for the user.

%%%%%%%%%%%%%%%%%%%%%%%%%%%%%%%%%%%%%%
\subsection{Indoor Reconstruction}
\label{SLAM}

% motivation: why 3d reconstruction: provide geometric accuracy
$3$D reconstruction is necessary to accurately understand the spatial relationships and structure of objects in a scene, as it transforms pixel-based $2$D data into a detailed $3$D representation. 
This process provides the depth and geometry needed for segmentation, where distinguishing between different parts of the scene is important, and for VR applications, where immersive and realistic representations of the environment are required. 

Recent studies in 3D reconstruction have favored Gaussian Splatting (GS) because it can represent the scene realistically and enhance camera localization~\cite{keetha:2024:CVPR:splatam,yan:2024:CVPR:gsslam,Matsuki:2024:CVPR:GSSLAM}. To perform $3$D reconstruction, where a sequence of $2$D RGB-D images is taken as input and the outputs include camera pose estimation and scene mapping, We adopt the state-of-the-art method SplaTAM~\cite{keetha:2024:CVPR:splatam}, followed by surface reconstruction to refine the results.

Although SplatAM estimates camera poses and 3D Gaussian scene, it mainly focues on speed and pose estimation. 
The quality of the rendered 3D Gaussians is not sufficient for VR applications, which require higher fidelity and precision.
To address this issue, we modified the PGSR method by introducing two geometric losses to refine the 3D scene.

In the surface reconstruction stage, we take the RGB-D images and camera poses estimated from the previous stage as input.
We train gaussians using the RGB and geometric losses from PGSR~\cite{chen2024pgsr}, along with two new losses: $\mathcal{L}_{normal}$ and $\mathcal{L}_{depth}$.

First, we leverage geometric foundation model Metric3D~\cite{yin2023metric} to estimate the normal of the frame $t$ as $N_t^e$.
Since the predicted normal may not be consistent across the video, we use its derivatives in normal loss $\mathcal{L}_{normal}$ to guide the surface reconstruction.
We compute the first-order derivative of the predicted normal map $N_t^e$ and the rendered normal map $N_t^r$,
with respect to the pixel coordinates, 
\begin{align}
    \dot{N_t^e}(p) &= \frac{\partial N_t^e(p)}{\partial p},\\
    \dot{N_t^r}(p) &= \frac{\partial N_t^r(p)}{\partial p},
\end{align}
where $p$ is the pixel location. 
We constrain the $\dot{N_t^r}$ from 3D Gaussian scene to be close to the predicted derivative $\dot{N_t^e}$ by the L1 norm,
\begin{align}
    \mathcal{L}_{normal} &= 1 - \frac{1}{|p|} \sum_p \|\dot{N_t^e}(p) - \dot{N_t^r}(p) \|_1,
\end{align}
where $|p|$ is the total number of pixels.

Second, we constrain the rendered depth $D_t^r$ to be close to the captured depth $D_t^{gt}$ using the L1 norm,
\begin{align}
    \mathcal{L}_{depth} &=  \frac{1}{|p|} \sum_p \|D_t^r(p) - D_t^{gt}(p) \|_1.
\end{align}
We minimize the following loss function to optimize the 3D reconstruction,
\begin{equation}
    \mathcal{L} = \mathcal{L}_{PGSR} + \lambda_N \mathcal{L}_{normal} + \lambda_D \mathcal{L}_{depth},
\end{equation}
where we set $\lambda_N=1$ and $\lambda_D=1.5$ in our experiments.

%%%%%%%%%%%%%%%%%%%%%%%%%%%%%%%%%%%%%%
\subsection{3D Instance Segmentation}
\label{softgroup}

% We compare various 3D instance segmentation approaches, including those applied to different map representations. 
% The GS-based algorithm often struggles to distinguish between foreground objects and the background, leading to mixed or incomplete segmentation results, while segmentation on point clouds is more robust.

% To overcome the lack of robust instance-level segmentation algorithms for GS representations, 
Point cloud segmentation has been well-studied for decades. 
It is robust enough to segment objects in a zero-shot video.
Hence, we convert our 3D scene into point clouds and apply point cloud-based instance segmentation.
During the conversion, we use a robust truncated signed distance field (TSDF)~\cite{zhou2013dense} method, ensuring that the point cloud inherits the point colors from the 3D Gaussians.

For the instance segmentation task, we adopt SoftGroup++\cite{vu2023scalable, vu2022softgroup}, trained on the extensive ScanNet++\cite{yeshwanthliu2023scannetpp} dataset, as it is the most robust method identified, capable of handling a wide variety of room types.
The advantages of SoftGroup++ include its scalability and ability to operate effectively in large-scale scenes, such as those exceeding $10 m^2$. It also accurately produces bounding boxes for segmented instances.

\begin{figure*}[t]
    \centering
    \includegraphics[width=\linewidth]{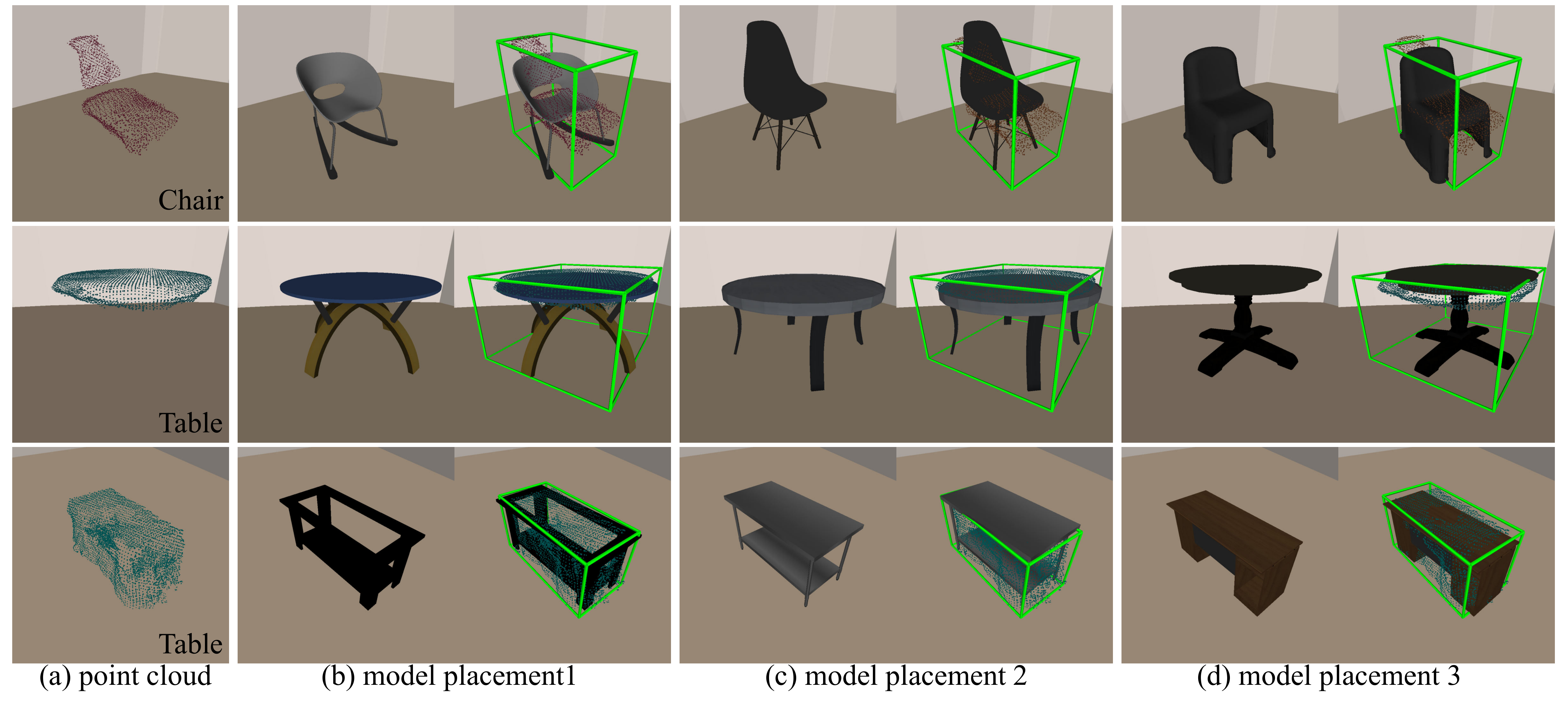}
    \caption{Results of the model placement in the 3D scene. We replace the segmented point cloud with the candidate models from the ShapeNet~\cite{chang2015shapenet} to explore various interior designs.
    Our model could randomly generate multiple model placements in the scene. We show three of them for qualitative evaluation. (a) is the segmented point cloud of the interested object. (b-d) show the model placements in the scene. The placement of furniture fit the point clouds well.}
    \label{fig:model-placement}
\end{figure*}

%%%%%%%%%%%%%%%%%%%%%%%%%%%%%%%%%%%%%%
\subsection{Envelope Extraction}
\label{dim-est}

We assume that the floor is a plane and a plane representation is used to model the floor.
Given the point cloud of the segmented floor, we ran a principal component analysis (PCA) to compute its main axes. 
The first and the second axes with the largest eigenvalues are denoted as the directional vectors of the floor, $f_1, f_2$.
The last axis is the normal of the floor $f_n = f_1 \times f_2 $.
We projected the point cloud of the floor to the 2D plane $f_1$ and $f_2$ and computed the 2D contour of them.
The floor area and dimensions are estimated from the 2D contour.

To reconstruct the wall in the scene, we projected the segmented wall point cloud to the normal of the floor $f_n$
The projected lengths are a list of the heights from the floor.
To remove the outliers and noises, we sort the heights and select the 95\% percentile number as the height of the room.
The volume of the room is computed by the reconstructed wall and floor.

\subsection{Model Retrieval and Registration}
\label{regis}

We use the widely-used ShapeNet~\cite{chang2015shapenet} as our 3D model database.
Given a point cloud of the query object and its label, we randomly picked $M$ 3D models from the same category as the model candidates.
The orientation of the 3D model in ShapeNet~\cite{chang2015shapenet} is pre-calibrated to face the z-axis, while the sizes and the rotations of xy-axes may not be aligned with the reconstructed scene.
Hence, we move the 3D object to the reconstructed scene and try to optimize its scales and rotation.

The baseline method is to directly place the 3D model from the ShapeNet in the scene that only aligns with the normal direction of the floor.
The potential problem with this straightforward way is that the placed model may not fit the scene well.
Specifically, the sizes of the model may not fit the segmented point cloud of the object of interest, and the orientations of the candidate model and the interested object may differ.

We evaluate the chamfer distance \emph{Dist} between the point cloud of the segmented object $T$ and the placed 3D model $M$. 
In practice, we estimate the major axes of $T$ and $M$, respectively. 
We first resize the 3D candidate model $M$ by the ratio of the lengths of the major axes.
Next, we iteratively rotate the object by small intervals (20 degrees) and then compute the chamfer distance \emph{Dist}.
We select the minimal \emph{Dist} as the object placement.

% The \emph{Axis Angle} measures the angles between the major axes of the point cloud and the 3D model, denoted as $A$.
% For the point cloud and the candidate model, we project them to the floor and use PCA to compute the major axes.
% Then we measure the average angle deviations of the major axes, respectively.

% In our experiment, to uniform the units of two metrics, we use the following score to estimate the quality of placement,
% \begin{equation}
%     S = Dist + \lambda A.
%     \label{eqn:quality}
% \end{equation}
% The smaller $S$ represents the better placement of the 3D model. 
% We set $\lambda=0.01$ in our experiments.

\section{Experiments}

We conducted extensive experiments on both the TUM dataset and our self-captured dataset to evaluate the performance of our automated pipeline.

For the self-captured dataset, we used an iPhone Pro with a LiDAR depth sensor to capture $200$ RGB-D images for each scene. We captured $10$ scenes, covering classrooms, library study rooms, and dormitory.
For the open dataset, we chose to use the TUM RGB-D dataset~\cite{tum2012benchmark}.
This is because none of our modules have seen the TUM dataset; it is a new scene compared to ScanNet++, on which SoftGroup++~\cite{vu2022softgroup} was trained.
These dataset provides a sufficient variety to demonstrate the robustness of our pipeline on unseen scenes.

We pick up 3D assets from ShapeNet~\cite{chang2015shapenet} as our 3D model database. 
These assets are then aligned with the layout to generate the interior design variations.

\subsection{Quantitative Results}

We evaluated the placement of the 3D objects using the proposed metrics.
Table~\ref{tab:furniture-placement} shows the comparisons between the baseline and our method.
The proposed metrics can effectively evaluate the quality of the model placement in the indoor design. 
Poor indoor design and incorrect model placement result in low scores.
Compared with the baseline, our method can align the model orientation with the segmented point clouds, as well as the sizes of the placed model also fit.

The results indicate that our method outperforms the baseline in both test scenes. Specifically, for the TUM-plant scene, our approach achieves a significantly lower distance error ($0.0189$ vs. $0.0538$), and similarly, in the Study Room scene, the error is reduced from $0.0181$ to $0.0070$. These results demonstrate the superior accuracy of our approach in furniture placement.
\begin{figure*}[t]
    \centering
    \includegraphics[width=\linewidth]{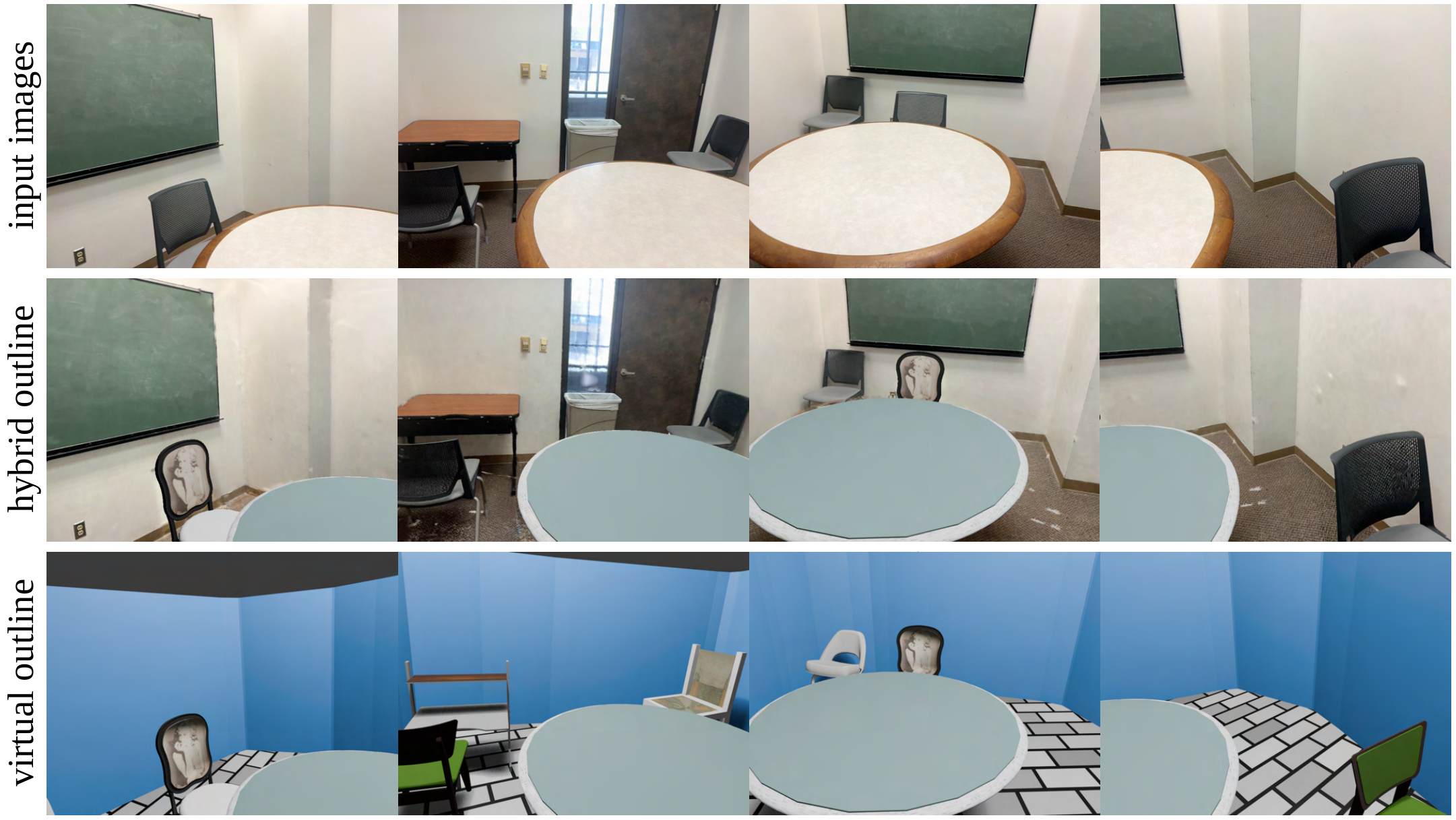}
    \caption{This illustration shows the design results from our automated pipeline. (Top row) The captured input image; (Middle row) The hybrid outline, which includes both the reconstructed gaussians and CAD models; (Bottom row) The virtual outline, where all elements are replaced by the CAD model.}
    \label{fig:results}
\end{figure*}

\subsection{Qualitative Results}

\begin{table}[t]
  \caption{Quantitative comparisons on the model placements. Our approach has better placement accuracy compared with the baseline method.}
  \label{tab:furniture-placement}
  \begin{tabular}{lccc}
    \toprule
    Scene & Metrics & Baseline & \textbf{Ours}\\
    \midrule
    \multirow{1}{*}{TUM-plant} & \textit{Dist}~$\downarrow$ & 0.0538 ($\pm 0.095$) & 0.0189 ($\pm 0.026$) \\
%    & \textit{Axis Angle}~$\downarrow$ & 33.06$^{\circ}$ ($\pm 7.23$) & 11.49$^{\circ}$ ($\pm 6.71$)\\
    \midrule
    \multirow{1}{*}{Study room} & \textit{Dist}~$\downarrow$ & 0.0181 ($\pm 0.018$) & 0.0070 ($\pm 0.005$) \\
 %   & \textit{Axis Angle}~$\downarrow$ & 26.40$^{\circ}$ ($\pm 7.66$) & 7.48$^{\circ}$ ($\pm 7.26$)\\
  \bottomrule
\end{tabular}
\end{table}

Figure~\ref{fig:model-placement} shows the qualitative results of our automatic indoor design, with three examples presented for qualitative evaluation.
We replace the segmented point cloud with the candidate models of ShapeNet~\cite{chang2015shapenet} to explore various interior designs.
From the results, we can conclude that our approach effectively adjusts both the scale and orientation of the 3D model to align with the point cloud, ensuring accurate placement in the scene. Furthermore, it is capable of generating multiple plausible design variations.

Figure~\ref{fig:results} illustrates the design results from our automated pipeline. 
It can be observed that the original position is accurately preserved, with no penetration into the walls or object overlap, ensuring spatial consistency and realistic layout placement, regardless of whether the hybrid or virtual outline option is used.

\section{Conclusion}
In this work, we introduce AURORA, a novel automated pipeline designed to generate both hybrid and virtual outlines from a single set of RGB-D images.
Our pipeline integrates advanced techniques in 3D reconstruction, segmentation, and model placement, enabling the generation of realistic and diverse interior designs. 
Through extensive experiments on both public datasets and self-captured data, we demonstrate the robustness and versatility of AURORA in handling a variety of scenes and configurations. 
The results show that our approach effectively preserves spatial consistency, handles complex environments, and generates plausible interior design variations, making it a powerful tool for both design professionals and automated design applications.

\section{Limitation}
Although the AURORA system integrates multiple methodologies into a cohesive framework, errors introduced in earlier stages can propagate and adversely impact subsequent stages.
As illustrated in Figure~\ref{fig:results}, the accuracy of furniture registration is influenced by the quality of the segmented point cloud.
In future work, we intend to investigate an end-to-end training approach to mitigate such issues and improve the overall performance.

\begin{acks}
This research is partly supported by Texas A\&M University ASCEND: Research Leadership Fellows Program.  
Yongqing Liang is partly supported by NSF CBET 2115405.
We thank Zixi Liu for preliminary experiments and data collection.
\end{acks}

\bibliographystyle{ACM-Reference-Format}
\bibliography{acmart}

\end{document}